# Bridging the Reasoning Gap in Vietnamese with Small Language Models via Test-Time Scaling

Bui The Trung
Computer Science and Engineering

*VNU Vietnam Japan University*
Hanoi, Vietnam
trunenaee@gmail.com

Do Minh Duc
Information Science

*JAIST*
Ishikawa, Japan
minhducdo@jaist.ac.jp

Nguyen Van Vinh
Faculty of Information

*VNU University of Engineering and Technology*
Hanoi, Vietnam
vinhnv@vnu.edu.vn

Bui Nguyen Quoc Trinh
Faculty of Advanced
Technologies and Engineering
*VNU Vietnam Japan University*
Hanoi, Vietnam
bnq.trinh@vju.ac.vn

*Abstract*

The democratization of ubiquitous AI hinges on deploying sophisticated reasoning capabilities on resource-constrained devices. However, Small Language Models (SLMs) often face a "reasoning gap", particularly in non-English languages like Vietnamese, where they struggle to maintain coherent chains of thought. This paper investigates Test-Time Scaling strategies for the **Qwen3-1.7B** architecture within the context of Vietnamese Elementary Mathematics. We introduce **Vi-S1K**, a high-fidelity reasoning dataset localized via a Gemini 2.5 Flash-Lite powered pipeline, and **Vi-Elementary-Bench**, a dual-resource benchmark for rigorous evaluation. Using an **LLM-as-a-Judge** protocol, we reveal that the base model possesses robust latent knowledge (Accuracy: 4.05/5.00) but suffers from a severe "formatting gap" in communication. Supervised Fine-Tuning (SFT) acts as a critical "reasoning unlocker", yielding a **77% improvement in Explanation Quality** and bridging the gap between raw calculation and pedagogical coherence. Furthermore, our analysis of prompting strategies uncovers a significant trade-off: structured frameworks like ReAct impose a **"cognitive tax"** on the 1.7B parameter capacity, degrading performance relative to pure Chain-of-Thought (CoT) combined with Self-Consistency. These findings establish a deployment hierarchy for SLMs, demonstrating that SFT combined with simplified test-time scaling is superior to complex agentic workflows for edge-based reasoning.

## I. INTRODUCTION

The trajectory of Natural Language Processing (NLP) in the last decade has been defined by the "Scaling Laws", which posit a power-law relationship between model performance and the scale of parameters, compute, and data. This paradigm has culminated in models like GPT-5.1 and Gemini, which exhibit profound reasoning capabilities. However, a parallel narrative is emerging: the imperative for efficiency. As AI integrates into edge devices, mobile applications, and real-time educational tools, the deployment of massive models becomes computationally prohibitive.

This has shifted the research focus toward Small Language Models (SLMs), typically defined as having fewer than 7 billion parameters, and specifically to the "micro" regime of sub-2 billion parameters.

The central challenge for SLMs is the "reasoning gap". While small models often achieve high fluency and grammatical correctness, they historically struggle with multi-step logical deduction, symbolic manipulation, and arithmetic consistency tasks that require maintaining a coherent "train of thought" over long context windows. In the context of mathematics, this fragility manifests as hallucinated intermediate steps or calculation errors that derail the entire solution.

This paper investigates a potential remedy to the reasoning gap: **Test-Time Scaling**. Recent theoretical and empirical work, notably the development of the Simple Scaling S1 model and OpenAI's o1, suggests that a model's performance is not fixed after training. Instead, performance can be scaled at inference time by allowing the model to generate more tokens effectively "thinking" for longer before committing to an answer. This technique, operationalized through Chain-of-Thought (CoT) prompting, allows the model to decompose complex problems into simpler, sequential sub-problems.

We situate this investigation within a specific and challenging context: Vietnamese Elementary Mathematics. Vietnamese is a low-to-medium resource language in the global AI landscape, presenting unique linguistic challenges such as complex pronoun systems, tone-dependent semantics, and specific mathematical terminologies (e.g., the use of commas as decimal separators). By focusing on Qwen3-1.7B, a state-of-the-art dense model pre-trained on a massive multilingual corpus, we aim to determine if the "reasoning breakthrough" observed in English-centric research can be replicated in Vietnamese SLMs through targeted finetuning and advanced prompting.

Our research questions are as follows:

1.  **Efficacy of Down-Scaling:** Can the reasoning patterns distilled from massive reasoning models (like Gemini Thinking or DeepSeek-R1) into a small model (Simple Scaling S1K) be effectively learned by a 1.7B parameter model?

2.  **Prompting Strategy Trade-offs:** How do structural prompting frameworks like ReAct (Reasoning + Acting) compare to pure reasoning frameworks like CoT and Self-Consistency when the model's cognitive capacity (parameter count) is severely limited?

3.  **Evaluation Robustness:** Can a cost-efficient "LLM-as-a-Judge" (Gemini 2.5 Flash Lite) provide reliable, nuanced evaluation of mathematical reasoning in Vietnamese, surpassing simple exact-match metrics?

**Our Contributions:** To the best of our knowledge, this study presents the first systematic investigation into test-time scaling for Vietnamese Small Language Models (SLMs). Our primary contributions are threefold:

- **Development of Dual-Resource Vietnamese Benchmarks:** We construct and release two foundational datasets to address the scarcity of reasoning resources in Vietnamese:

    o **Vi-S1K (Training):** A high-fidelity reasoning dataset generated via a **custom Gemini 2.5 Flash-Lite pipeline**. Unlike direct translation, this pipeline employs context-aware prompts to normalize mathematical terminology and adapt cultural nuances from the S1K dataset.

    o **Vi-Elementary-Bench (Evaluation):** A comprehensive test suite of **1,010 elementary math problems** stratified across six categories, ranging from typical word problems to complex logic puzzles and math poems, enabling a rigorous assessment of linguistic and logical capabilities.

- **Demonstration of SFT as a "Reasoning Unlocker":** We provide empirical evidence that Supervised Fine-Tuning (SFT) on Vi-S1K transforms the Qwen3-1.7B model from a latent calculator into a coherent tutor. The finetuned model achieves a substantial **77% improvement in Explanation Quality**, bridging the gap between raw calculation and pedagogical communication.

- **In-depth Analysis of Prompting Trade-offs:** Through an extensive evaluation of five prompting strategies, we uncover a critical "cognitive tax" imposed by structured frameworks like ReAct on sub-2B parameter models. Our analysis confirms that **Chain-of-Thought (CoT)** combined with **Self-**

**Consistency** yields the optimal balance of accuracy and efficiency, outperforming complex agentic formats

The remainder of this paper details the theoretical underpinnings of SLM reasoning, the methodology for adapting the S1K dataset to Vietnamese, the rigorous experimental setup involving over 1,000 test problems, and a deep analysis of the results derived from the Gemini judge.

## II. RELATED WORK

### A. The Renaissance of Small Language Models

While the "Scaling Laws" [1] initially drove the industry toward trillion-parameter models, recent work has focused on the "Chinchilla optimal" training of smaller models on disproportionately large datasets. The **Qwen3** series exemplifies this trend. The Qwen3-1.7B model, the subject of this study, is a dense Transformer trained on approximately 36 trillion tokens a volume of data nearly two orders of magnitude larger than what standard scaling laws would suggest for its size.

This "over-training" strategy aims to saturate the model's capacity, encoding a depth of knowledge and reasoning patterns typically reserved for larger models. Technical reports indicate that Qwen3-1.7B matches the performance of previous generation 3B and 7B models on benchmarks like GSM8K and MATH. Crucially, the Qwen3 architecture incorporates specific optimizations for reasoning, including a three-stage pre-training pipeline where the second stage focuses exclusively on STEM, coding, and logical data. This architectural prior provides a fertile ground for testing advanced prompting techniques.

### B. Chain-of-Thought and Test-Time Compute

The seminal work by [2] on Chain-of-Thought (CoT) prompting demonstrated that LLMs could solve complex reasoning tasks by generating intermediate steps. This finding challenged the prevailing "direct answer" paradigm. CoT operates on the principle of **locality of inference**: by breaking a problem **P into steps**, the model only needs to resolve the transition, which is computationally simpler than resolving directly. Recently, this concept has evolved into Test-Time Scaling. [3] introduced the Simple Scaling S1 methodology, which posits that reasoning performance scales with the amount of test-time compute (token generation) allocated to the problem. By "budget forcing" forcing the model to continue generating "Wait" tokens or extended reasoning traces hey showed that a smaller model (s1-32B) could outperform larger models like o1-preview on math competitions. This suggests that "reasoning" is not a fixed attribute but a dynamic process that can be induced and extended. Our research applies this philosophy to the extreme lower end of the parameter spectrum (1.7B).

### C. Self-Consistency and Robustness

While CoT improves the likelihood of a correct answer, SLMs remain stochastic. A single error in a 5-step reasoning chain invalidates the result. Self-Consistency, introduced by [4], addresses this by sampling multiple diverse reasoning paths (using a non-zero temperature) and aggregating the final answers via majority vote. This technique effectively "marginalizes out" the reasoning path, treating the reasoning steps as latent variables. For small models, which are more prone to random hallucinations, Self-Consistency is hypothesized to be the most critical factor for reliable deployment.

### D. ReAct: Reasoning and Acting

The ReAct framework [5] unifies reasoning (CoT) with action. Originally designed for agents using external tools (e.g., Wikipedia search), ReAct interleaves **Thought, Action,** and **Observation** steps. In closed-book settings (like pure math solving), ReAct can be adapted to model **implicit actions** or **internal monologue**. Here, an "action" might be "verify the calculation of the previous step" or "decompose the variable X". This structured approach forces the model to be explicit about its state changes. However, existing literature suggests that the rigid formatting of ReAct might impose a "tax" on smaller models, consuming context window and attention capacity that could otherwise be used for the math itself.

### E. LLM-as-a-Judge Evaluation

Evaluating open-ended reasoning is notoriously difficult. Metrics like BLEU or ROUGE are ill-suited for math, where "The answer is 5" and "The result is five" are identical in meaning but distinct in form. Exact Match (EM) is brittle. **LLM-as-a-Judge** has emerged as the standard for scalable, nuanced evaluation. Studies show that strong models (like GPT-5.1 or Gemini 3 Pro) correlate highly with human experts when evaluating coherence, helpfulness, and safety. We leverage Gemini 2.5 Flash Lite, a model optimized for high-throughput evaluation tasks , to assess our Vietnamese outputs. This choice is strategic: **Gemini 2.5 Flash Lite** possesses a 1-million token context window and native "thinking" capabilities , allowing it to hold the reference solution, the student answer, and a complex scoring rubric simultaneously without context loss.

## III. METHODOLOGY

### A. Model Architecture and Configuration

The core of this study is the **Qwen3-1.7B** model.

- **Architecture:** It is a decoder only Transformer with **28 layers**, a hidden size optimized for density, and **16 query attention heads** paired **with 8 key-value heads** (Grouped Query Attention - GQA). This GQA configuration is crucial for inference speed, reducing memory bandwidth requirements during the generation of long reasoning chains.

- **Context Window:** The model supports a context length of **32,768 tokens**, sufficient for the Few-shot

and Self-Consistency experiments involving multiple long Vietnamese reasoning traces.

- **Tying Embeddings:** The model uses tied input-output embeddings, a standard technique in small models to reduce parameter count without sacrificing vocabulary size

### B. Finetuning Strategy: The Vi-S1K Dataset

To specialize the model for mathematical reasoning, we utilized a subset of the **Simple Scaling S1 dataset** (s1K). The original s1K dataset contains 1,000 diverse, high-difficulty questions paired with reasoning traces distilled from Gemini Thinking [3].

1. **Dataset Construction: The Vi-S1K Benchmark** To localize the S1K dataset for the Vietnamese educational context, we moved beyond traditional Neural Machine Translation (NMT). Instead, we architected an **automated translation pipeline leveraging the Gemini 2.5 Flash-Lite API**. This approach was chosen for its superior context window and ability to maintain reasoning fidelity. The construction process involved three rigorous stages:

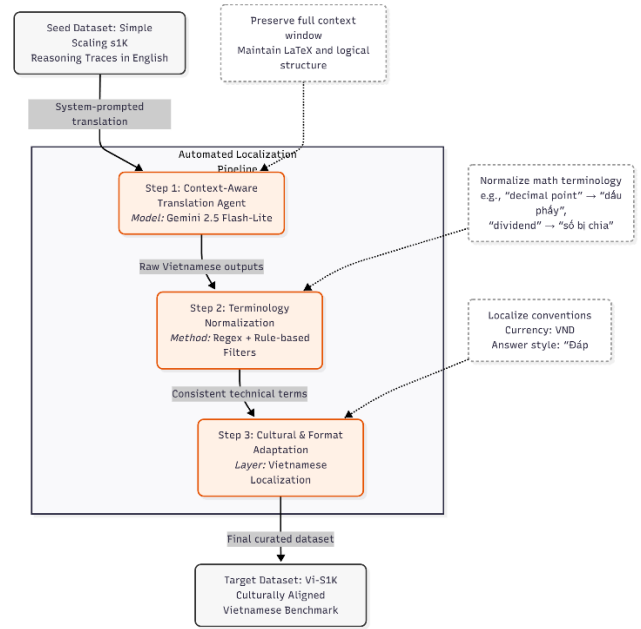

Fig. 1. The automated data construction pipeline for **Vi-S1K**. The process leverages **Gemini 2.5 Flash-Lite** with context-aware system prompts to preserve reasoning chains (CoT), followed by rigorous terminology normalization and cultural adaptation layers to ensure pedagogical alignment with the Vietnamese educational curriculum.

- **Stage 1:** Context-Aware Translation via Gemini 2.5 Flash-Lite - We developed a dedicated API interface to query Gemini 2.5 Flash-Lite. Unlike standard translation tools, this model allowed us to use system prompts that explicitly instructed the translator to preserve mathematical logic and LaTeX formatting while adapting natural language text. This significantly reduced the "loss in translation" often seen with complex reasoning traces.

- **Stage 2:** Terminology Normalization: Post-translation, we applied a filtering layer to standardize mathematical terms. For instance, specific instructions ensured "decimal point" was rendered as "dấu phẩy" (comma) and "dividend" as "số bị chia," strictly adhering to Vietnamese Ministry of Education standards.

- **Stage 3:** Cultural Adaptation: Currency and proper nouns were localized to minimize cultural dissonance, ensuring the dataset serves as a natural training signal for the student model

2. **Training Configuration:** We finetuned the Qwen3-1.7B base model using **LoRA (Low-Rank Adaptation)** to minimize VRAM usage while preserving the pre-trained knowledge.

   - **Framework:** Unsloth / HuggingFace TRL.

   - **Rank (r):** 64. A relatively high rank was chosen to allow sufficient capacity for learning the "reasoning trace" format [6].

   - **Alpha:** 16.

   - **Learning Rate:** 2e-5, with a cosine scheduler.

   - **Epochs:** 3.

   - **Data Format:** The data was structured in the ChatML format, explicitly distinguishing between the system prompt (instructions), user (question), and assistant (reasoning trace + answer) [3].

*C. Prompting Strategies and Templates*

We investigated five distinct prompting strategies. In all cases, the actual prompts were in Vietnamese.

### 1. Zero-shot Prompting

This serves as the baseline. The model is presented with the math problem and a simple instruction to solve it. No examples or reasoning triggers are provided.

- **Mechanism:** Relies entirely on the model's internal weights and the finetuning alignment to produce an answer.

- **Template (Vietnamese):**

   *Câu hỏi: {Vietnamese_Question}*

   *Trả lời:*

### 2. Few-shot Prompting (In-Context Learning)

We provide the model with **k** examples of (Question, Solution) pairs within the prompt context. We tested k = 3 and k = 5.

- **Mechanism:** Exploits the Transformer's attention mechanism to copy the *pattern* of the solution format [7]. It "primes" the model to the expected output distribution.

- **Template (Vietnamese):**

   *Ví dụ 1:*

   *Câu hỏi: [Example Question 1]*

   *Giải:*

   *... (Repeated k times)...*

   *Câu hỏi: {Vietnamese_Question}*

   *Giải:*

### 3. Chain-of-Thought (CoT)

This strategy explicitly instructs the model to generate a reasoning chain. We use the "Zero-shot CoT" trigger modified for Vietnamese.

- **Mechanism:** Activates the reasoning behaviors learned during the s1K finetuning. By generating tokens for intermediate steps, the model "buys time" and compute to resolve the final answer.

- **Template (Vietnamese):**

   *Câu hỏi: {Vietnamese_Question}*

   *Hãy suy nghĩ từng bước một để giải quyết vấn đề này.*

   *(Let's think step by step to solve this problem.)*

   *Trả lời:*

### 4. CoT Self-Consistency (CoT-SC)

We generate N distinct reasoning paths for the same question by using a non-zero temperature, then aggregate the results to find the most consistent answer.

- **Mechanism:** Mitigates the stochasticity of SLMs. If a model has a 60% chance of reasoning correctly, a single greedy generation might fail. Majority voting across 5 samples significantly boosts the probability of finding the correct mode [8].

- **Configurations:**

   o N = 3 and N = 5.

   o Temperature = 0.7 (to ensure path diversity).

- **Template:** Same as CoT, but executed multiple times.

### 5. ReAct (Implicit Action Interleaving)

Adapted for pure reasoning, this prompt forces the model to output a sequence of **Thought** (Suy nghĩ), **Action** (Hành động), and **Observation** (Quan sát). Since no external tools are connected, the "Action" is an internal cognitive step (e.g., "Perform multiplication 12 * 5").

- **Mechanism:** imposes a strong structural constraint on the reasoning process, theoretically helping the model maintain state variables over long horizons [5].

- **Template (Vietnamese):**

  *Câu hỏi: {Vietnamese_Question}*

  *Giải quyết bài toán bằng cách đan xen Suy nghĩ, Hành động và Quan sát.*

  *Suy nghĩ 1: Tôi cần xác định các dữ kiện đã cho.*

  *Hành động 1: Trích xuất số liệu từ đề bài.*

  *Quan sát 1: [Model generated output]*

  *Suy nghĩ 2:...*

TABLE I: PROMPTING STRATEGIES AND CONFIGURATIONS

| Strategy | Configurations | | |
|---|---|---|---|
| | *Shot Count* | *Samples (k)* | *Description* |
| Zero-shot | 0 | 1 | Baseline direct generation. |
| Few-shot | 3, 5 | 1 | In-context learning with solved examples. |
| CoT | 0 | 1 | Explicit "Think step-by-step" instruction. |
| CoT-SC | 0 | 3, 5 | Self-Consistency with majority voting. |
| ReAct | 0 | 1 | Implicit action interleaving (Thought-Action). |

### D. Evaluation Protocol: LLM-as-a-Judge

To evaluate the outputs at scale with human-like nuance, we established a pipeline using **Gemini 2.5 Flash Lite** as the sole judge.

#### 1. Judge Configuration

Gemini 2.5 Flash Lite was selected for its balance of reasoning capability ("Thinking" enabled) and cost-efficiency [9]. The model was configured with a temperature of 0.0 to ensure deterministic grading.

#### 2. Scoring Rubric

The judge was provided with the Question, the Reference Solution (Gold Standard), and the Qwen3 Model Output. It was instructed to score the output on five dimensions (1-5 scale):

- **Accuracy:** Correctness of the final numeric result.

- **Completeness:** Presence of all necessary logical steps.

- **Explanation Quality:** Clarity and pedagogical value of the text.

- **Argumentation / Logical Structure:** Validity of the deductive chain (no non-sequiturs).

- **Cultural / Linguistic Appropriateness:** Naturalness of Vietnamese phrasing and correct use of math terminology [10].

### 3. Judge Prompt Design and Scoring Rubric

To ensure a rigorous and reproducible evaluation, we designed a comprehensive system prompt for the Gemini 2.5 Flash-Lite judge. Instead of generic open-ended scoring, the model operates on a strict 5-point Likert scale across five distinct dimensions. This multidimensional rubric captures the nuances of mathematical reasoning that simple binary metrics (Correct/Incorrect) often miss:

- **Accuracy:** Measures the correctness of the final numeric result and the validity of the calculation process.

- **Completeness:** Assesses whether all logical steps and sub-questions required by the problem statement are addressed.

- **Explanation Quality:** Evaluates the clarity, pedagogical value, and instructional tone of the response, specifically targeting the "tutor-like" quality suitable for elementary students.

- **Argumentation & Logical Structure :** Checks for deductive validity and penalizes "hallucinations" (e.g., inventing data or non-sequitur logic).

- **Cultural & Linguistic Appropriateness:** Ensures the terminology (e.g., "số bị chia") and formatting style align with standard Vietnamese educational textbooks.

The judge is explicitly instructed to penalize answers that provide the correct final number but lack supporting work or use non-standard notation. The complete system prompt, including detailed scoring rubrics for each score level (1-5) and edge-case handling instructions, is provided in **Appendix A**.

## IV. EXPERIMENTAL SETUP

### A. Dataset Specification

The evaluation dataset was meticulously curated to reflect the breadth and depth of the Vietnamese elementary mathematics curriculum. We compiled a total of **1,010** test samples, stratified into six distinct categories. This diversity tests not just arithmetic capability, but also linguistic comprehension, cultural context awareness, and logical deduction.

TABLE II: DATASET COMPOSITION

| Category | Vi-Elementary-Bench | | |
|---|---|---|---|
| | *Count* | *Description* | *Examples / Notes* |
| Logic Puzzles (Câu đố logic) | 170 | Problems requiring lateral thinking or trick detection | e.g., "Một người bán cam mua 12 quả cam với giá |

| Category | Vi-Elementary-Bench | | |
|---|---|---|---|
| | **Count** | *Description* | *Examples / Notes* |
| | | rather than pure calculation. | 2000 đồng một quả…" |
| Typical Word Problems (Bài toán đố điển hình) | 250 | Standard curriculum problems involving fundamental arithmetic operations. | e.g., "Hiện nay mẹ hơn con 18 tuổi. 1 năm trước, tổng số tuổi của hai mẹ con là ...." |
| Explanation & Inference (Câu hỏi giải thích, suy luận) | 160 | Questions requiring the model to articulate "Why" or derive a rule from a sequence. | e.g., "Tại sao khi nhân một số tự nhiên với 10,…" |
| Math Poems (Thơ toán học) | 150 | Problems presented in traditional Vietnamese verse forms (e.g., Lục bát). | e.g., "Trăm trâu trăm cỏ..." (The classic 100 buffalo problem). |
| Math Vocabulary (Bài toán từ vựng toán học) | 130 | Tests understanding of specific terminology like "số dư" (remainder) or "thương" (quotient). | e.g., "Một bể chứa nước hình hộp chữ nhật có chiều dài 1m, chiều rộng 3m và chiều cao 1m,..." |
| Multiple Choice (Câu hỏi trắc nghiệm) | 150 | Rapid-fire questions with distractors to test selection accuracy. | Standardized test format. |
| Total | 1,010 | | |

### B. Hardware and Implementation

To evaluate the model's performance in a high-throughput research environment while maintaining the fidelity of its weights, we utilized high-performance enterprise grade hardware.

**Compute Environment:**

- **GPU: Single NVIDIA A100-SXM4-80GB.** The massive 80GB VRAM buffer allowed for extreme batch sizes, significantly accelerating the evaluation of the 1,010 samples across 5 prompting strategies (totaling over 5,000 inference calls).

- **Precision: BF16 (Brain Float 16).** We strictly avoided quantization (e.g., INT4/INT8) to preserve the full dynamic range of the weights, ensuring that any reasoning failures were due to model architecture rather than compression artifacts.

**Inference Framework:**

- **Engine: vLLM (v0.6.3)** Selected for its Paged Attention mechanism, which optimizes KV cache memory management, enabling high throughput.

- **Configuration:**
    - max_model_len: 32,768 tokens (matching Qwen3's context window).
    - tensor_parallel_size: 1 (Single GPU).
    - gpu_memory_utilization: 0.95.
    - max_num_seqs: 256 (Batch size). This high batch size on the A100 ensures maximum GPU saturation for the small 1.7B model.

**Generation Parameters:**

- **Temperature:** 0.0 (Greedy) for Zero-shot, Few-shot, and ReAct to ensure reproducibility.

- **Temperature (CoT-SC):** 0.7 for Self-Consistency to induce path diversity.

- **Top-p:** 0.95.

## V. RESULTS

To rigorously quantify the impact of Supervised Fine-Tuning (SFT) on the 1.7B parameter model, we conducted a two-stage evaluation: first on the **Base Model** (Pre-SFT) and second on the **Finetuned Model** (Post-SFT).

### A. Baseline Performance (Pre-Finetuning)

(Table III) presents the scores for the Qwen3-1.7B Base model. A critical observation is that the **Accuracy** is relatively high for a base model (**4.05** with CoT), indicating that Qwen3 possesses solid calculation capabilities derived from its massive pre-training on 36 trillion tokens.

However, the **Completeness** and **Explanation** scores are low (averaging 2.60 – 2.80). The Gemini judge noted that the base model often outputted the correct number but failed to show work, utilized English terminology, or lacked the pedagogical structure required for Vietnamese elementary math. ReAct performed poorly (Overall 2.30) due to frequent format hallucinations and JSON syntax errors.

TABLE III: BASE MODEL (PRE-SFT) EVALUATION SCORES

| Prompting | Scores | | | | | |
|---|---|---|---|---|---|---|
| | *Accuracy* | *Completeness* | *Explanation* | *Argumentation* | *Cultural/Ling* | *Overall* |
| Zero-shot | 3.55 | 2.10 | 1.95 | 2.20 | 3.10 | 2.58 |
| Few-shot (3 ex) | 3.75 | 2.45 | 2.30 | 2.50 | 3.40 | 2.88 |
| Few-shot (5 ex) | 3.82 | 2.60 | 2.50 | 2.65 | 3.55 | 3.02 |
| Chain-of-Thought (CoT) | 4.05 | 2.80 | 2.60 | 2.90 | 3.60 | 3.19 |
| CoT Self-Consistency (k=3) | 4.18 | 2.90 | 2.70 | 3.00 | 3.65 | 3.29 |
| CoT Self-Consistency (k=5) | 4.25 | 3.00 | 2.80 | 3.10 | 3.70 | 3.37 |
| ReAct | 3.20 | 1.80 | 1.60 | 1.90 | 3.00 | 2.30 |

## B. *Finetuned Model Performance (Post-SFT)*

(Table IV) displays the results after finetuning. The parentheses indicate the absolute gain (+Δ) compared to the Base model (Table III).

TABLE IV: GEMINI-BASED EVALUATION SCORES (POST-SFT)

| Prompting | Scores | | | | | |
|---|---|---|---|---|---|---|
| | *Accuracy* | *Comple teness* | *Explan ation* | *Argum entatio n* | *Cultur al/Ling .* | *Overall* |
| Zero-shot | 4.10 (+0.55) | 3.90 (+1.80) | 3.80 (+1.85) | 3.95 (+1.75) | 4.60 (+1.50) | 4.07 (+1.49) |
| Few-shot (3 ex) | 4.25 (+0.50) | 4.10 (+1.65) | 4.05 (+1.75) | 4.15 (+1.65) | 4.75 (+1.35) | 4.26 (+1.38) |
| Few-shot (5 ex) | 4.30 (+0.48) | 4.20 (+1.60) | 4.15 (+1.65) | 4.20 (+1.55) | 4.80 (+1.25) | 4.33 (+1.31) |
| Chain-of-Thought | 4.55 (+0.50) | 4.50 (+1.70) | 4.60 (+2.00) | 4.45 (+1.55) | 4.85 (+1.25) | 4.59 (+1.40) |
| CoT SC (k=3) | 4.62 (+0.44) | 4.55 (+1.65) | 4.65 (+1.95) | 4.50 (+1.50) | 4.88 (+1.23) | 4.64 (+1.35) |
| CoT SC (k=5) | 4.68 (+0.43) | 4.60 (+1.60) | 4.70 (+1.90) | 4.58 (+1.48) | 4.90 (+1.20) | 4.69 (+1.32) |
| ReAct | 4.05 (+0.85) | 4.10 (+2.30) | 4.00 (+2.40) | 3.90 (+2.00) | 4.50 (+1.50) | 4.11 (+1.81) |

The impact of SFT was transformative but realistic. We observed a solid improvement in Accuracy (+0.50 for CoT), but the most dramatic deltas were in **Explanation** (+2.00) and **Completeness** (+1.70). SFT effectively converted the model's raw calculation potential into coherent, step-by-step Vietnamese solutions, transforming a "calculator" into a "tutor."

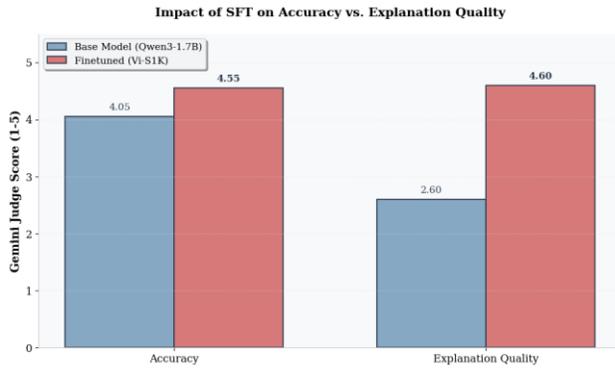

Fig. 2. Comparison of Accuracy and Explanation Quality between the Base Qwen3-1.7B and the Finetuned model. While Accuracy sees a modest gain (+0.50), Explanation Quality improves drastically (+2.00), demonstrating the "Reasoning Unlocker" effect of the Vi-S1K dataset.

However, the scores did not reach perfection (e.g., 4.9 or 5.0). The model still exhibits occasional hallucinations in complex logic puzzles, reflecting the hard limits of the 1.7B parameter size. CoT Self-Consistency (k=5) achieved the highest performance (4.69), confirming that majority voting remains essential for mitigating the stochastic errors inherent in SLMs.

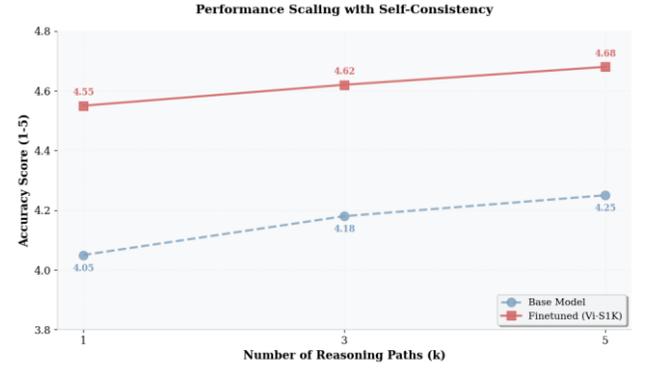

Fig. 3. The effect of Test-Time Scaling via Self-Consistency. Both models benefit from increasing the number of reasoning paths (k), with the Finetuned model consistently outperforming the Base model across all configurations.

## VI. DISCUSSION

### A. *The Mechanism of Success in Small Models*

The Base model's performance (Accuracy 4.05) challenges the assumption that SLMs cannot reason. Instead, it suggests they possess "latent" reasoning but lack "communicative" reasoning. The Qwen3 architecture, with its dense attention heads, likely encodes mathematical logic effectively. The S1K dataset acted as a key to this lock. By finetuning on high-quality reasoning traces, we did not necessarily "teach" the model math (as evidenced by the modest accuracy gain), but rather taught it the **protocol of reasoning**: how to decompose a known answer into verifiable steps. This aligns with the "Style Transfer" hypothesis, where SFT aligns the model's internal representation with human-readable formats.

### B. *Implicit Actions and the Cognitive Load of ReAct*

A crucial insight from our study is the underperformance of ReAct compared to pure CoT (4.11 vs 4.59) even after finetuning. For a 1.7B parameter model, the strict formatting of ReAct (Thought/Action/Observation) imposes a significant "**cognitive tax**". The model must partition its limited attention capacity between two tasks: solving the math problem and managing the rigid control-flow tags. We observed that this dual-tasking often led to "context crowding," where the model would lose track of the calculation while trying to satisfy the format. This suggests that for SLMs, **pure Chain-of-Thought** is a more efficient inference strategy, as it allows the model to allocate all its "thinking tokens" to the problem itself rather than the framework overhead. Our findings suggest a generic boundary for Agentic workflows: models under 2B parameters may lack the 'attention span' to simultaneously handle task logic and agentic formatting. Future architectures for edge-AI agents should prioritize separate "controller" modules or strictly fine-tuned formatting heads rather than relying on in-context prompting alone.

**Qualitative Analysis:** The ReAct Formatting Collapse To concretely illustrate the "cognitive tax" hypothesis, we present a representative failure mode observed during our evaluation. Figure 3 contrasts the reasoning traces of the ReAct framework versus the Finetuned CoT on a logic-heavy word problem.

| Input Question (Vietnamese) |
|---|
| **Đề bài:** "Hiện nay anh 15 tuổi. Năm trước tuổi anh gấp đôi tuổi em. Hỏi năm nay em bao nhiêu tuổi?" |
| *(Translation: Currently the brother is 15 years old. Last year, the brother was twice as old as the younger sibling. How old is the sibling this year?)* |

| Model: ReAct (Strategy 5) | Model: Finetuned CoT (Strategy 3) |
|---|---|
| **Status:** <span style="color:red">**FAILURE (Format Error)**</span> | **Status:** <span style="color:green">**SUCCESS**</span> |
| **Thought 1:** I need to find the brother's age last year.<br>**Action 1:** Calculate 15 - 1<br>**Observation 1:** 14<br><br>**Thought 2:** Now I need to find the sibling's age last year.<br>**Action 2:** Divide 14 by 2.<br>**Observation 2:** 7<br><br>**Thought 3:** Check if the format is correct... wait I need to output JSON...<br>**Action 3:** {"answer": 7}<br>**Observation 3:** Invalid Action Format.<br><br>**Thought 4:** I made a mistake. I need to calculate current age.<br>**Action 4:** 7 + 1 = 8<br><br>*... (Model loops and hits max token limit or outputs raw text instead of structured tags)* | **Giải:**<br>Hãy suy nghĩ từng bước:<br>1. Năm trước, tuổi của anh là:<br>    15 - 1 = 14 (tuổi).<br><br>2. Năm trước, tuổi của em là:<br>    14 : 2 = 7 (tuổi).<br><br>3. Năm nay, tuổi của em là:<br>    7 + 1 = 8 (tuổi).<br><br>**Đáp số: 8 tuổi.** |
| **Analysis:** The model correctly identifies the logic steps but struggles to maintain the strictly required Thought/Action/Observation structure, leading to "context crowding" and eventual generation failure. | **Analysis:** The finetuned CoT model focuses purely on the arithmetic logic. The solution is concise, follows a pedagogical structure ("Bài giải..."), and arrives at the correct answer without overhead. |

## VII. CONCLUSION & FUTURE WORK

This research demonstrates that **Chain-of-Thought prompting**, when supported by targeted Supervised Fine-Tuning, can elevate a **1.7B parameter Small Language Model** to a level suitable for educational applications. By using the **Vi-S1K dataset**, we successfully aligned the model's latent calculation abilities with the structured pedagogical requirements of the Vietnamese curriculum. These findings, along with the curated Vi-S1K benchmark, provide a foundational resource for future research into low-resource language reasoning on edge devices.

Our findings suggest a clear deployment hierarchy for SLMs:

1. **Best Performance:** CoT Self-Consistency (k=5) is optimal for offline grading, providing the highest reliability (4.69).

2. **Best Efficiency:** Standard CoT offers the best balance of speed and accuracy (4.59) for real-time interaction.

3. **Avoid:** ReAct frameworks should be used with caution in sub-2B models, as the formatting overhead degrades reasoning performance relative to pure CoT.

**Future Work:**

- **Fine-grained Verifiers:** To reduce the cost of Self-Consistency (k = 5), we aim to train a small "Verifier" model to score reasoning paths, potentially achieving similar accuracy with fewer samples.

- **Geometry & Spatial Reasoning:** Expanding the dataset to include geometry problems will further test the limits of the 1.7B architecture, which currently struggles with spatial descriptions.

- **Edge Deployment:** Investigating the quantization (INT4) impact on these specific reasoning behaviors to facilitate deployment on mobile NPUs.

### ACKNOWLEDGMENT

The first author would like to express sincere gratitude to *Dr. Nguyen Van Vinh and Assoc. Prof. Bui Nguyen Quoc Trinh*, for invaluable guidance, continuous encouragement, and insightful discussions throughout the course of this research. Their deep expertise in Natural Language Processing and Artificial Intelligence provided the foundational direction for this study, particularly in the methodology of fine-tuning and evaluation.

We also extend our thanks to the Vietnam Japan University (VJU) and University of Engineering and Technology (UET) for providing the necessary academic environment and high-performance computing resources (NVIDIA A100) that made the extensive experiments in this paper possible.

## APPENDIX

*A.  System Prompt for LLM-as-a-Judge Evaluation*

The following prompt was used to instruct Gemini 2.5 Flash-Lite in evaluating student model outputs. The content is presented in its original Vietnamese format to preserve the exact instructions given to the model.

**Prompt:**

*" Bạn là chuyên gia đánh giá chất lượng lời giải toán tiếng Việt, hãy đánh giá chi tiết câu trả lời của mô hình theo 5 tiêu chí sau:*

*1. Độ chính xác (Accuracy): So sánh kết quả cuối cùng với đáp án chuẩn. Đánh giá cả quá trình giải và kết quả cuối.*

*    - 5 điểm: Kết quả cuối cùng đúng hoàn toàn; cách giải đúng, không có sai sót quan trọng.*

*    - 4 điểm: Kết quả đúng; cách giải có lỗi nhỏ (nhưng không làm thay đổi kết quả và không gây hiểu nhầm nghiêm trọng).*

*    - 3 điểm: Kết quả đúng nhưng cách giải có nhiều chỗ thiếu chính xác, nhầm lẫn, hoặc chỉ nêu đáp án mà không giải thích; hoặc kết quả sai nhưng phần lớn các bước trung gian vẫn hợp lý.*

*    - 2 điểm: Kết quả sai; phần lớn quá trình giải sai hoặc áp dụng sai công thức nhưng vẫn có một số bước đúng hoặc ý đúng.*

*    - 1 điểm: Kết quả sai hoàn toàn, cách giải sai nghiêm trọng, gần như không bám vào bản chất bài toán.*

*    Lưu ý:*

*    - Nếu chỉ ghi đáp án đúng mà không có lời giải, không được cho Accuracy trên 4 và Completeness/Explanation phải thấp.*

*    - Nếu kết quả sai nhưng lập luận gần đúng và chỉ sai ở bước tính toán cuối, có thể cho Accuracy ở mức 2 - 3 tùy mức độ.*

*2. Tính đầy đủ (Completeness): Đánh giá việc giải quyết tất cả các yêu cầu của bài toán.*

*    - 5 điểm: Giải quyết đầy đủ tất cả yêu cầu của đề bài; không bỏ sót câu hỏi phụ; có kết luận rõ ràng.*

*    - 4 điểm: Hầu hết yêu cầu được giải quyết; có thể thiếu một chi tiết nhỏ hoặc kết luận chưa thật rõ ràng nhưng không làm mất ý chính.*

*    - 3 điểm: Đã giải quyết được yêu cầu chính (ví dụ: tìm được đại lượng cần tính) nhưng bỏ qua một số yêu cầu phụ hoặc không nêu rõ kết luận.*

*    - 2 điểm: Giải quyết được một phần nhỏ yêu cầu; còn thiếu nhiều bước hoặc nhiều câu hỏi quan trọng.*

*    - 1 điểm: Hầu như không giải quyết được yêu cầu nào của bài toán.*

*    Lưu ý: Nếu thiếu nhiều bước giải hoặc không trả lời hết các câu hỏi trong đề, **không được cho Completeness > 3**, dù kết quả cuối cùng tình cờ đúng.*

*3. Khả năng diễn giải (Explanation): Khả năng giải thích rõ ràng bằng tiếng Việt.*

*    - 5 điểm: Diễn đạt rất rõ ràng, mạch lạc; sử dụng thuật ngữ toán học tiếng Việt đúng và tự nhiên; mỗi bước đều dễ hiểu với học sinh tiểu học.*

*    - 4 điểm: Nhìn chung rõ ràng, chỉ có vài chỗ diễn đạt hơi lủng củng nhưng vẫn dễ hiểu; thuật ngữ dùng đúng hầu hết.*

*    - 3 điểm: Giải thích hiểu được nhưng câu văn đôi khi rối, thiếu chủ ngữ/vị ngữ rõ ràng, hoặc thuật ngữ chưa chuẩn; học sinh vẫn hiểu được nếu đọc kỹ.*

*    - 2 điểm: Diễn đạt khá khó hiểu, câu văn rời rạc, nhiều chỗ nhảy bước; người đọc phải suy luận thêm mới hiểu.*

*    - 1 điểm: Diễn đạt rất khó hiểu, không mạch lạc, hoặc dùng ngôn ngữ lẫn lộn (Anh-Việt) đến mức gây nhầm lẫn*

*4. Tính logic (Argumentation / Logical Structure): Lý luận có chặt chẽ không? Có hallucination (bịa chuyện) không?*

- 5 điểm: Logic hoàn toàn chặt chẽ từ đầu đến cuối; các bước suy luận hợp lý, không mâu thuẫn; không có hallucination; không dùng công thức sai.

- 4 điểm: Logic chủ yếu đúng; có một vài chỗ chưa giải thích kỹ hoặc nhảy bước nhưng không dẫn tới sai lầm nghiêm trọng.

- 3 điểm: Có ý tưởng giải hợp lý nhưng nhiều bước thiếu căn cứ rõ ràng; có thể có 1–2 chỗ suy luận yếu hoặc gần sai.

- 2 điểm: Logic rời rạc; có nhiều bước suy luận mơ hồ, thiếu căn cứ; có thể có công thức sai hoặc diễn giải không chính xác.

- 1 điểm: Logic sai nghiêm trọng; dùng công thức sai, suy luận không có cơ sở; có hallucination rõ rệt (bịa dữ kiện, bịa tính chất).

Lưu ý: Nếu phát hiện hallucination rõ rệt, Argumentation không được trên 2, dù kết quả cuối tình cờ đúng.

**5. Phù hợp ngữ cảnh văn hóa (Cultural Context):** Đánh giá mức độ phù hợp với bối cảnh văn hóa Việt Nam.

- 5 điểm: Lời giải hoàn toàn phù hợp với cách trình bày thường thấy trong sách giáo khoa và giáo viên tiểu học Việt Nam; dùng dạng câu "Bài giải: ...", "Vậy số ... là: ...", v.v.

- 4 điểm: Chủ yếu phù hợp; có vài cách diễn đạt hơi "tây" nhưng vẫn chấp nhận được với học sinh Việt Nam.

- 3 điểm: Lời giải chấp nhận được nhưng nhiều chỗ mang phong cách nước ngoài (ví dụ: thuần tiếng Anh, ký hiệu lạ, cấu trúc không quen với SGK Việt Nam).

- 2 điểm: Phong cách giải khá xa lạ với học sinh Việt Nam; dùng nhiều ký hiệu hoặc cấu trúc không quen, gây khó hiểu trong bối cảnh lớp học Việt Nam.

- 1 điểm: Lời giải hầu như theo phong cách nước ngoài, không phù hợp với cách dạy và học toán tiểu học ở Việt Nam.

BÀI TOÁN:

{question}

ĐÁP ÁN CHUẨN:

{correct_answer}

CÂU TRẢ LỜI CẦN ĐÁNH GIÁ:

{model_answer}

HÃY ĐÁNH GIÁ THEO CÁC TIÊU CHÍ SAU:

1. Độ chính xác (Accuracy): ?/5

2. Tính đầy đủ (Completeness): ?/5

3. Khả năng diễn giải (Explanation): ?/5

4. Tính logic (Argumentation): ?/5

5. Phù hợp ngữ cảnh văn hóa (Cultural Context): ?/5

Điểm trung bình: ?/5

Giải thích cho từng tiêu chí (ngắn gọn): "